\pdfoutput=1

\documentclass[11pt]{article}

\usepackage{acl}

\usepackage{times}
\usepackage{latexsym}

\usepackage[T1]{fontenc}
\usepackage{makecell}
\usepackage{booktabs}
\usepackage{multirow}

\usepackage[utf8]{inputenc}
\usepackage{amsmath}
\usepackage{amssymb}
\usepackage{verbatim}

\usepackage{microtype}
\usepackage{caption}
\usepackage{subcaption}
\usepackage{graphicx}
\usepackage{dsfont}

\title{WeCheck: Strong  Factual Consistency Checker via Weakly Supervised Learning}

\author{ \parbox{\linewidth}{\centering Wenhao Wu\textsuperscript{1}\thanks{\ \ Work is done during an internship at Baidu Inc.}, Wei Li\textsuperscript{2}, Xinyan Xiao\textsuperscript{2}, Jiachen Liu\textsuperscript{2}, Sujian Li\textsuperscript{1}\thanks{\ \ Corresponding author.}, Yajuan Lyu\textsuperscript{2}}\\
    \textsuperscript{1}Key Laboratory of Computational Linguistics, MOE, Peking University \\
  \textsuperscript{2}Baidu Inc., Beijing, China \\
  \texttt{\{waynewu,lisujian\}@pku.edu.cn}\\
  \texttt{\{liwei85,xiaoxinyan,liujiachen,lvyajuan\}@baidu.com}\\}

\begin{document}
\maketitle
\begin{abstract}
A crucial issue of current text generation models is that they often uncontrollably generate text that is factually inconsistent with inputs.
Due to lack of annotated data, existing factual consistency metrics usually train evaluation models on synthetic texts or directly transfer from other related tasks, such as question answering (QA) and natural language inference (NLI).
Bias in synthetic text or upstream tasks makes them perform poorly on text actually generated by language models, especially for general evaluation for various tasks.
To alleviate this problem, we propose a weakly supervised framework named \textbf{WeCheck} that is directly trained on actual generated samples from language models with weakly annotated labels.
WeCheck first utilizes a generative model to infer the factual labels of generated samples by aggregating weak labels from multiple resources.
Next, we train a simple noise-aware classification model as the target metric using the inferred weakly supervised information.
Comprehensive experiments on various tasks demonstrate the strong performance of WeCheck, achieving an average absolute improvement of $3.3\%$ on the TRUE benchmark over 11B state-of-the-art methods using only 435M parameters.
Furthermore, it is up to $30\times$ faster than previous evaluation methods, greatly improving the accuracy and efficiency of factual consistency evaluation.\footnote{Our metric can be easily accessed from \url{https://huggingface.co/nightdessert/WeCheck}} 
\end{abstract}

\section{Introduction}
The research of text generation has achieved significant progress in recent years, but it still suffers the main issue of generating output which is factually inconsistent with the given inputs~\cite{maynez-etal-2020-faithfulness}.
To tackle this issue, various metrics have been  designed to check the consistency between  generated text and the given inputs~\cite{kryscinski-etal-2020-evaluating,scialom-etal-2021-questeval}.
As we know, how to construct  such a  metric has attracted increasing attention in a variety of fields~\cite{wu-etal-2022-precisely}, including text summarization~\cite{kryscinski-etal-2020-evaluating,wu-etal-2022-frsum}, dialogue generation~\cite{welleck-etal-2019-dialogue}, and text simplification~\cite{devaraj-etal-2022-evaluating}.

\begin{figure}
\centering
\includegraphics[scale=0.3]{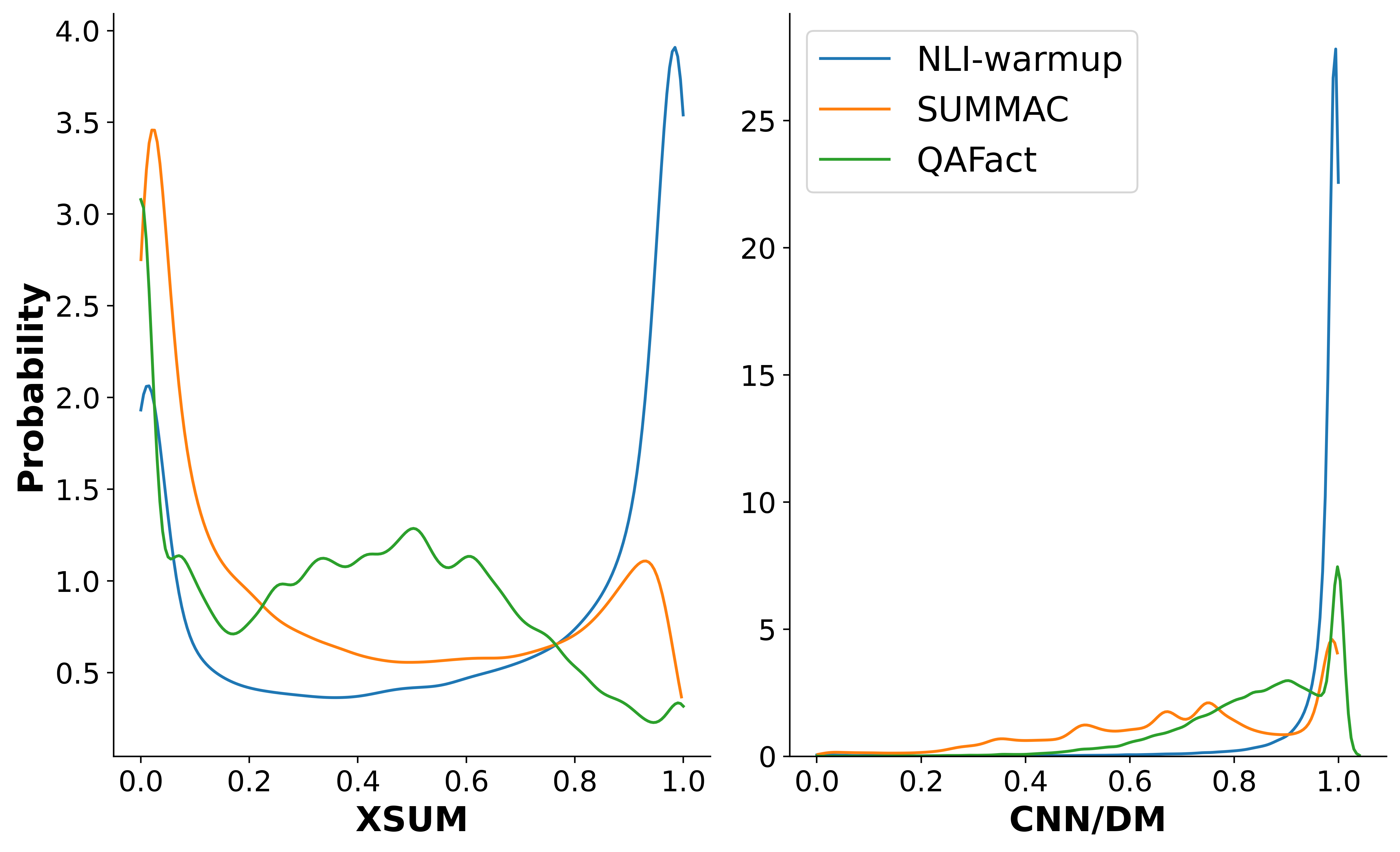}
\caption{Probability density of   factual consistency   scores predicted by different metrics sampled from  BART on  XSum and CNN/DM datasets. The horizontal axis is the score ranged in [0, 1], and the vertical axis is the probability density.  }
    \label{fig:prob_density}
\end{figure}

Existing factual metrics can be classified into two types: one based on synthetic data and the other based on task transfer.
Synthetic-data based metrics~\cite{kryscinski-etal-2020-evaluating, mishra-etal-2021-looking} apply data augmentation techniques to construct factual and non-factual texts as positive and negative samples, respectively.
Metrics trained from these synthetic samples often perform poorly due to the significant mismatch between features of actual generated and synthetic text (e.g. distribution of factual errors) ~\cite{goyal-durrett-2021-annotating}.
Task-transfer based metrics utilize the reasoning ability of models trained on relevant upstream tasks, such as natural language inference (NLI) ~\cite{falke-etal-2019-ranking, laban-etal-2022-summac} and question answering (QA) ~\cite{wang-etal-2020-asking, fabbri-etal-2022-qafacteval} and directly apply them to evaluate factual consistency without any adaption.

As described above, previous metrics are learned indirectly from other related 
resources but without seeing the actual generated text.
In such cases, they may overfit to their upstream tasks and fail to generalize to actual generated samples that have significantly different data features.
Figure~\ref{fig:prob_density} illustrates the probability density of three  metrics, where the horizontal axis  is metric scores and the vertical axis is the score density.  
Though these metrics are comparable in performance, they vary significantly in probability distributions,
especially in the XSUM dataset, where sample features are greatly different from upstream tasks of these metrics\footnote{In XSum, the summary of  each document is 
abstractive, while existing NLI and QA datasets do not have this feature.}, NLI-warmup is extremely confident in predicting both very high and low scores while SUMMAC and QAFact are only confident in predicting low scores\footnote{For more details about these metrics please refer to \S~\ref{sec:nli_warm} and \S\ref{sec:baseline}.}.
Furthermore, during testing, ensembling different metric scores by simply averaging will further improve their performance~\cite{honovich-etal-2022-true}.
This also implies that the evaluation metrics learned from different resources are also complementary.

To bridge the gap between training and testing and mitigate the scarcity of labeled data, in this paper, we propose \textbf{WeCheck}, a factual consistency \textbf{Check}ing framework based on \textbf{We}akly supervised learning.
Specifically,
WeCheck is based on a learning paradigm that provides weak supervision via modeling multiple label sources without access to ground truth.
Different from previous metrics,  WeCheck directly utilizes the abundant actual generated samples bootstrapped from models trained on target downstream tasks, e.g. BART on text summarization.
Then, WeCheck follows a two-step pipeline consisting of weak annotation and noise-aware fine-tuning to get the target metric model.

In the weak annotation step, by aggregating multiple weak supervision resources, we infer the unknown ground truth label of a sample. 
To reach this goal, we first provide each sample with a set of weak supervision signals calculated from various other metrics.
These metrics are learned from various resources or tasks such as  QA-based metrics and NLI-based metrics.
After unifying and filtering these signals, we train a generative labeling model that models agreements and disagreements between them to infer the likelihood of their latent ground truth label.
The inferred ground truth  likelihood is then treated as a probabilistic label to provide weak supervision.
In the second step, we apply noise-aware fine-tuning to train the target metric model.
It is noted here, the weak annotation also brings noises to the supervision signal and brings new challenges to the model optimization process.
As a solution, we first warmup our target metric model with NLI data for a better initialization before weakly supervised training.
Then, after filtering out samples that are likely to be noisy, we finetune our target metric model with weak annotations.
In summary, WeCheck could learn how to utilize multiple resources for weak annotation while recognizing and filtering the potential noises accompanied by weak supervision.

Experimental results show that WeCheck not only achieves state-of-the-art performance but also is computationally efficient.
On the TRUE benchmark~\cite{honovich-etal-2022-true}, which is the current most comprehensive benchmark for factual consistency evaluation, WeCheck obtains an average ROC AUC of 84.8, 3.3\% absolute improvement over previous 11B pre-trained task transferred metrics with only a size of 435M parameters.
Moreover, it's much more stable for various generation tasks, with much lower variance on different tasks.
Thus, WeCheck is a simple but more effective and efficient metric for factual consistency evaluation.

We summarize our contributions as follows:
\begin{itemize}
\item We propose a novel  factual consistency evaluation metric based on weakly supervised learning, namely WeCheck, which  is 
directly trained on actual generated samples from
language models with weakly annotated labels.
\item WeCheck is both effective and  efficient achieving $3.3\%$ absolute improvement and up to $30$ times faster comparing with  previous state-of-art metrics. 

\item WeCheck is a general metric which is  also more stable on various generation tasks  and datasets than previous methods.
\end{itemize}

\begin{figure}
    \centering
    \includegraphics[scale=0.44]{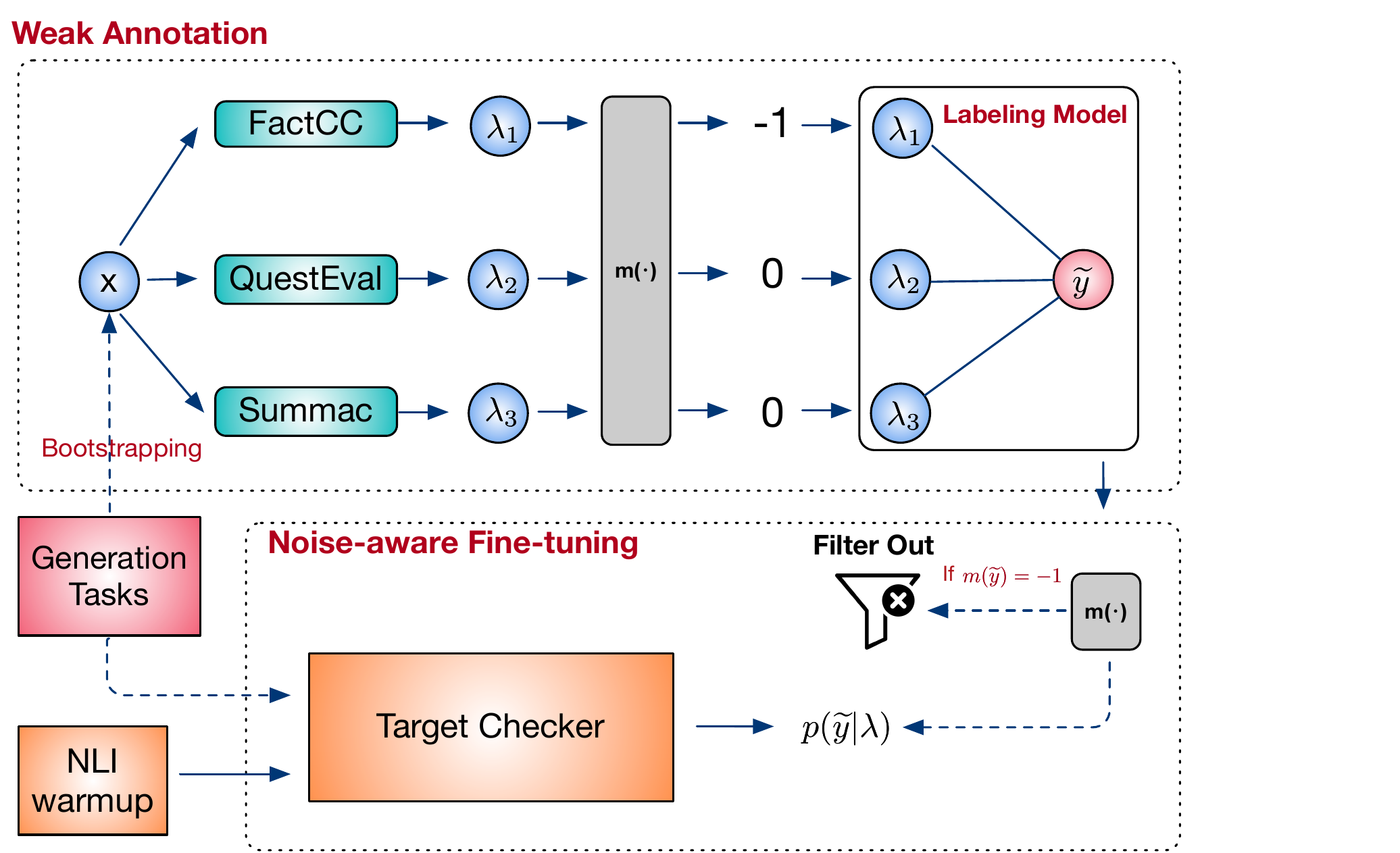}
    \caption{The overall framework of WeCheck,  including weak annotation and noise-aware fine-tuning.
    Weak annotation infers the likelihood of each sample's true label based on its weak supervision signal set $\boldsymbol{\lambda}$, and noise-aware fine-tuning trains the target metric with the inferred likelihood of ground-truth label.
    }
    \label{fig:wecheck_framework}
\end{figure}

\section{WeCheck Framework}\label{sec:framework}
Figure~\ref{fig:wecheck_framework} illustrates the two-step  pipeline of WeCheck  framework.
In the upper part of the figure, during the weak annotation step, we first calculate a set of weak supervision signals for each sample bootstrapped from target generation tasks.
Then, we use a mapping function to unify the weak supervision signals and infer the likelihood of the ground truth label of each sample.
After annotation, we apply noise-aware fine-tuning to train our target metric model, shown in the lower part of the figure.
Noise-aware fine-tuning first warmup target metric model with NLI data and training it with filtered probabilistic labels.
In the following, we introduce our problem definition and detailed method.

\subsection{Problem Definition}
\paragraph{Factual Consistency Evaluation} 
Given a textual sequence as a premise, and another textual sequence as a hypothesis, which may be a generated summary or dialogue, the goal of a factual consistency metric $f_{\theta}$ is to predict whether the hypothesis is factual consistent given the premise.
For simplicity, we follow the previous textual entailment based framework~\cite{DBLP:conf/emnlp/KryscinskiKMXS19}, which takes $\boldsymbol{x}$, the concatenation of hypothesis and premise, as the input format and unifies the evaluation as a binary classification problem: $f_{\theta}(\boldsymbol{x}) \in [0, 1]$, where the predicted logit indicates the probability of  $\boldsymbol{x}$ being factually consistent.
Another advantage of using the entailment-based framework is that it is effective in terms of time complexity compared with other methods~\cite{laban-etal-2022-summac}.
Taking $f_{\theta}$ as the target metric model, the goal of WeCheck is to train $f_{\theta}$  into an efficient factual consistency metric. 

\paragraph{Weakly Supervised Training} 
In our weakly supervised settings, we first bootstrap a set of samples from the generation tasks, e.g. text summarization, and dialogue generation.
Using various factual metrics trained from multiple resources, we provide each sample  $\boldsymbol{x}$ with a set of weak signals $\boldsymbol{\lambda}  = (\lambda_1, \dots, \lambda_k )$, where each $\lambda_i$ is a logit separately calculated by a metric.
We treat the ground truth label $\widetilde{y}$ of $\boldsymbol{x}$ as a hidden variable that can be estimated by aggregating $\boldsymbol{\lambda}$.
To reach this goal, we train a labeling model $p_{\phi}$ to model agreements and disagreements relations between weak signals in $\boldsymbol{\lambda}$ and estimate the probability distribution  of the truth label,   $p_{\phi}(\widetilde{y}|\boldsymbol{\lambda})$.
Then, we apply $p_{\phi}(\widetilde{y}|\boldsymbol{\lambda})$ to supervise the metric model $f_{\theta}$.

\subsection{Weak Annotation}
To provide weak supervision for training, we follow data programming~\cite{ratner2017snorkel, DBLP:conf/icml/BachHRR17},  a weakly supervised learning paradigm based on modeling multiple label sources.
However, in data programming, weak supervision signals are often produced by various checking clauses, e.g. \textit{whether word “causes” appears in the sentence ?} and produce a discrete weak signal $\lambda_i \in \{0, 1, -1\}$,  where 0/1 stands for a vote for positive/negative label and $-1$ stands for a abstain vote.
However, in our scenario, due to the diversity of  metric frameworks, outputs of different metrics often do not share a unified output format and are usually continuous.
For example, QA-based metrics often produce continuous logits in $\left[0, 1\right]$, and NLI-based metrics often produce discrete labels of entailment or contradiction. 
Thus, the first thing before training the labeling model is to unify weak supervision signals by a mapping function, $ m\left(\lambda_i\right) \rightarrow \{0, 1, -1\} $.
In this way, we can model the transformed $\boldsymbol{\lambda}$ by a data programming based labeling model.
\paragraph{Weak Signal Unification}
We first unify all the weak supervision signals from different metrics into the same format,  a  logit $\lambda_i \in [0, 1]$. 
For the metric with single logit output, we directly use its output as $\lambda_i$.
For multi-label classification output, we select the probability of predicting entailment.
Notice that all the signals predicted by imperfect metrics will introduce a portion of noises. 
For a more reliable signal, the core idea for designing a mapping function $m$ is to map signals that the metric has high confidence into $\{0, 1\}$ and abstain low-confidence signals by mapping them to $-1$. 
Generally, this can be achieved by setting thresholds on signals. 
But another important issue to be noticed is that, as shown in Figure~\ref{fig:prob_density}, signal distributions vary significantly across metrics and datasets, which makes threshold selection difficult. 
Thus, we instead dynamically determine thresholds by setting constant probability mass  that contains the highest  confidence.
Specifically, we choose to map the lowest $p^-$ percent and the highest  $p^+$ percent of signal scores into label 0 and 1, separately, and map the rest interval of low-confident scores into -1.
Given the inverse cumulative distribution function of the $i$-th signal $F_i$,  we can calculate its positive and negative threshold $\gamma^+_i$ and $\gamma^-_i$ by:
\begin{equation}
   \gamma_i^+=F_i(1-p^+), \quad  \gamma_i^-=F_i(p^-)
.\end{equation}
The mapping function is then defined by:
\begin{equation}
    m(\lambda_{i}) = \left \{\begin{array}{rcl}
         0 & &\lambda_{i} \leq \gamma^-_i\\
         1 & &\lambda_{i} \geq \gamma^+_i\\
         -1& & \gamma^-_i <\lambda_{i} < \gamma^+_i.    \end{array}\right.
    \label{eq:mapping_func}
\end{equation}
For simplicity, we share $p^-$ and $p^+$ across different resources and datasets.
By applying the mapping function, we unify each $\lambda_i$ into a discrete label in $\{0,1,-1\}$.

\paragraph{Labeling model} We treat the true label $\widetilde{y}$ of  $\boldsymbol{x}$ as a  hidden variable and train the labeling model $p_\phi$ to estimate $\widetilde{y}$  by aggregating $\boldsymbol{\lambda}$\footnote{All the weak supervision signals in $\boldsymbol{\lambda}$ have already been converted into discrete labels by the mapping function $m$.}.
The generative model $p_\phi$ models the generation process of $\boldsymbol{\lambda}$ and  $\widetilde{y}$ by their joint probability.
Because all the weak supervision signals are inferred from different resources, we treat them as independent variables.
Then, given the prior $p(\widetilde{y})$\footnote{$p(\widetilde{y})$ usually depends on class distribution in a dataset. For simplicity, we set it as a uniform distribution.}, the joint probability is formulated by 
\begin{equation}
     p_{\phi}(\boldsymbol{\lambda},\widetilde{y})  = \prod_{\lambda_i \in \boldsymbol{\lambda}} p_{\phi}(\lambda_i,\widetilde{y})=\prod_{\lambda_i \in \boldsymbol{\lambda}} p\left(\lambda_{i}|\widetilde{y}\right) p\left(\widetilde{y}\right), \label{eq:joint_probability}
\end{equation}
following Bayesian rule.
Next, we need to model the likelihood $p\left(\lambda_{i}|\widetilde{y}\right)$ that labels the sample with $\lambda_i$ based on the latent label $\widetilde{y}$.
Following~\cite{ratner2017snorkel}, we define the labeling process of $\lambda_i$  as a sequence of Bernoulli process. 
Concretely, the $i$-th metric  has a probability of  $\beta_i$ not to abstain the sample and a probability $\alpha_i$ to label it correctly. 
Then, we calculate the likelihood by
\begin{equation}
    p_{\phi}(\lambda_{i}|\widetilde{y}) = \left \{\begin{array}{rcl}
         \beta_i\alpha_i & &\lambda_{i}\neq-1 \wedge \lambda_{i}=\widetilde{y}\\
         \beta_i(1 - \alpha_i) & &\lambda_{i}\neq-1 \wedge \lambda_{i}\neq \widetilde{y}\\
          1-\beta_i & & \lambda_{i}=-1,
    \end{array}\right. 
\end{equation}
where $\alpha_i, \beta_i$ are learnable hyper-parameters.
Given all samples, we train the labeling model by optimizing:
\begin{equation}
    \mathcal{L}_{\phi} = \min_{\phi} \sum_{\boldsymbol{\lambda}} \sum_{\widetilde{y} \in \{0, 1\}} \log p_{\phi} (\boldsymbol{\lambda},\widetilde{y}).
\end{equation}
\subsection{Noise Aware Fine-tuning}\label{sec:nli_warm}
\paragraph{NLI Warmup}
After we get the labeling model $p_\phi$, the next step is to train our metric model $f_{\theta}$ with the weak supervision inferred by it.
But in practice, we find direct training with weak supervision will cause the model easily converges to the local minima.
This may because reasoning over a long range of context is challenging and weak supervisions are also potential to be noisy.
These problems cause great difficulties in optimization.
Inspired by the idea of curriculum  learning~\cite{DBLP:conf/icml/BengioLCW09}, we first warmup our metric model on  NLI, an easier and closely related task.
We use the  mixture of four NLI datasets, MultiNLI  \cite{williams-etal-2018-broad}, Fever-NLI \cite{thorne-etal-2018-fever}, LingNLI~\cite{DBLP:conf/emnlp/ParrishHALNWAAL21} and Adversarial-NLI ~\cite{nie-etal-2020-adversarial}. 
Based on the warmed-up checkpoint, our metric model achieves much better results under weak supervision, which we will later show in our experiments. 

\paragraph{Noise Filtering and Training}
After warming up, we train our  metric model with weak supervision.
Because the estimated latent labels $\widetilde{y}$ can still be noisy due to the imperfect labeling model and weak supervision signals, we apply the likelihood of $\widetilde{y}$ that contains the certainty of the prediction as a soft probabilistic label  instead of the discrete label for training.
Based on the definition of joint probability in Eq.~\ref{eq:joint_probability}, we predict the likelihood of each sample by
\begin{equation}
     p_\phi(\widetilde{y}=1|\boldsymbol{\lambda}) = \frac{ p_\phi(\boldsymbol{\lambda},1)}{p_\phi(\boldsymbol{\lambda},1) + p_\phi(\boldsymbol{\lambda},0)}\, .\label{eq:soft_label}
\end{equation}
With convenience, we abbreviate $p_\phi(\widetilde{y}=1|\boldsymbol{\lambda})$  as $p(y^+)$.
Before training with $p(y^+)$, we first filter out estimated samples with low confidence, by applying the similar procedure in weak signal unification.
By reusing mapping function $m$, we filter out the low confident  probabilistic label and get the final training set by
\begin{equation}
    \mathcal{X} = \left \{ \left(\boldsymbol{x},p(y^+)\right) \middle|   m\left(p(y^+)\right)\neq-1 \right \},\label{eq:sample_filter}
\end{equation}
where $p(y^+)$  is the corresponding  probabilistic label of $\boldsymbol{x}$.
Then, given $f_\theta$ after warming up, we finetune  it by  
\begin{equation}
    \begin{aligned}
        \mathcal{L}_{f} = &\min_{\theta} \sum_{\boldsymbol{x} \in \mathcal{X}} \left[ p(y^+)\log\left(f_\theta(\boldsymbol{x})\right) \right.\\
        & + \left. (1-p(y^+)) \log(1-f_{\theta}(\boldsymbol{x})) \right],
    \end{aligned}
\end{equation}
where $p(y^+)$ is kept fixed without gradient back-propagation to  $p_\phi$ during training.

During inference, the model only needs to take the textual sequence $\boldsymbol{x}$ as input  and output the logit prediction $f_{\theta}(\boldsymbol{x})$.

\section{Experimental Settings}
In this section, we introduce the experimental settings of  WeCheck including the evaluation benchmark, baseline models, and implementation details.  
\subsection{TRUE Benchmark}
Recent works point out that the performance of a metric should be evaluated comprehensively across multiple tasks and datasets to reduce variance.
Thus, we evaluate WeCheck on TRUE~\cite{honovich-etal-2022-true}, a benchmark consisting of 11 datasets of 4 tasks including text summarization, dialogue generation, paraphrasing, and fact checking, where each sample in datasets is annotated with a binary label manually.
We only test on the first three tasks as fact checking is beyond our scope.
Following TRUE, we normalize each metric score into a logit  and report their  \textbf{Characteristic Area Under the Curve (ROC AUC)} w.r.t binary logits.
Evaluation with ROC AUC does not require metrics to set specific decision thresholds.
Details of tasks and datasets of TRUE   are introduce in the Appendix~\ref{sec:true_bench}.

\begin{table*}[htbp]
 \centering 
 \small
 \renewcommand\arraystretch{1.2}
 \setlength\tabcolsep{5.5pt}
 \begin{tabular}{l cccccc| c c c c| c|c c}\toprule
 & \multicolumn{6}{c}{\textbf{Summarization}} & \multicolumn{4}{c}{\textbf{Dialogue}}&\multicolumn{1}{c}{\textbf{Para.}}&\multirow{2}{*}{\textbf{Ave}}&\multirow{2}{*}{\textbf{Var}$\downarrow$}\\

 &Frank& SumE&MNBM&Q-C&Q-X&Ave&BEGIN&$Q^2$&DialF&Ave&PAWS\\
 \midrule
 BERTS
    &84.3&77.2&62.8&69.1&49.5&68.6&\textbf{87.9}&70.0&64.2&74.0&77.5&71.4&140\\
BARTS& 86.1&73.5&60.9&80.9&53.8&71.0&\underline{86.3}&64.9&65.6&72.3&77.5&72.2&132\\
 FactCC&
 76.4&75.9&59.4&76.4&64.9&70.6&64.4&63.7&55.3&61.1&64.0&66.7&60.1\\
 SC\textsc{zs}&
 \textbf{88.9}&\textbf{81.3}&71.1&80.9&78.1&80.1&82.0&77.4&84.1&81.2&\underline{88.2}&\underline{81.4}&30.4\\
 QuestEval&84.0&70.1&65.3&64.2&56.3&68.0&84.1&72.2&77.3&77.9&77.3&71.4&87.7\\
 QAFact&87.8&77.4&68.7&\underline{83.3}&76.9&\underline{78.8}&76.3&80.4&84.5&80.4&85.0&80.0&34.4\\
 \midrule
   \multicolumn{13}{c}{\textbf{11B Large Models}}\\
    \midrule
 $Q^2$ &87.8&78.8&68.7&83.5&70.9&77.9&79.7&80.9&86.1&82.2&89.7&80.7&51.6\\
 ANLI &89.4&80.5&77.9&82.1&83.8&82.5&82.6&72.7&77.7&77.7&86.4&81.5&24.9\\

 \midrule
  \multicolumn{13}{c}{\textbf{Our Models}}\\
 \midrule

 NLI-warmup&
 85.7&73.7&\underline{73.5}&73.2&\underline{80.1}&77.2&80.5&\underline{83.5}&\underline{87.3}&\underline{83.8}&85.4&80.3&31.8\\
  WeCheck&\underline{88.1}&\underline{79.8}&\textbf{83.0}&\textbf{82.6}&\textbf{81.4}&\textbf{83.0}&84.6&\textbf{84.0}&\textbf{90.0}&\textbf{86.2}&\textbf{89.6}&\textbf{84.8}&\textbf{13.2}\\

 \bottomrule
 \end{tabular}
 \caption{ROC AUC scores of all baseline metrics on three evaluation tasks on TRUE benchmark, where \textit{Para.}, \textit{Q-C}, \textit{Q-X}  are the abbreviations of paraphrase,  QAGS-CNN/DM and QAGS-XSUM, respectively. \textit{Ave} in block and penultimate column indicate the average performance on each task and  the average performance on the overall benchmark, respectively. \textit{Var} indicates variance across datasets. Results in bold and in underline indicate the best and second best performance (not including 11B baselines, as our model only have 435M parameters that comparable with other baselines).}
 \label{tab:main_exp}
\end{table*}

\subsection{Baseline}\label{sec:baseline}
We evaluate  WeCheck by  comparing with recently proposed metrics. 
We categorize these baselines by types of their  methods.
\paragraph{NLI-based Metrics}  
\textbf{FactCC}~\cite{kryscinski-etal-2020-evaluating} is a BERT-based metric with synthetic training samples constructed from rule-based data augmentation.
\textbf{SUMMAC(\textsc{SCzs})}~\cite{laban-etal-2022-summac} aggregates sentence-level entailment scores for the final factual consistency score.
We only report the zero-shot version \textsc{SCzs} instead of the supervised version \textsc{SCconv} because it is more effective on the TRUE benchmark.
\textbf{ANLI}~\cite{honovich-etal-2022-true} directly apply a  large 11B T5 trained on Adversarial-NLI ~\cite{nie-etal-2020-adversarial} dataset for fact checking and achieve SOTA performance on TRUE.

\paragraph{QA-QG based Metrics}  \textbf{QuestEval}~\cite{scialom-etal-2021-questeval} is a QA-QG based metric that jointly measures factual consistency and semantic relevance, where the importance of  generated questions are weighted by a trained model.
\textbf{QAFactEval (QAFact)} ~\cite{fabbri-etal-2022-qafacteval} is a  metric designed by carefully optimizing each component of the QG-QA  framework.
$\textbf{Q}^2$, from the version of ~\citet{honovich-etal-2022-true}, replace all the component of QA-QG framework into  T5 11B large models.

\paragraph{Other Types}
\textbf{BERTScore (BERTS)} ~\cite{DBLP:journals/corr/abs-1904-09675} measure  the similarity of a generated text and its reference by aggregating token-level similarities of their contextual representations.
\textbf{BARTScore (BARTS)}~\cite{DBLP:journals/corr/abs-2106-11520}  evaluate the quality of generated text by its modeling perplexity of a fine-tuned BART~\cite{lewis-etal-2020-bart}.

\subsection{Implementation Details}
All the baseline metrics are tested based on their open-sourced codes.
The  metric model of WeCheck is based on powerful pre-trained language model DeBERTaV3~\cite{DBLP:journals/corr/abs-2111-09543}.
Following the description in \S~\ref{sec:framework}, we first warm up DeBERTaV3 on NLI datasets and apply it for weak supervised training.
As regards to  training data, we sample text summarization examples from BART fine-tuned on CNN/DM and XSum datasets. 
We sample dialogue generation examples from MemNet~\cite{DBLP:journals/corr/abs-1811-01241} and dodecaDialogue~\cite{shuster-etal-2020-dialogue} trained on WoW dataset following~\citet{honovich-etal-2021-q2}.
For paraphrase, we directly use samples in PAWS since it can be regard as  a consistency checking dataset itself. 
For weak signals, we apply QAFact ~\cite{fabbri-etal-2022-qafacteval}, SUMMAC~\cite{laban-etal-2022-summac}, and the NLI warmed up DeBERTaV3 (NLI-warmup) as to provide weak signals for each sample as default.
For weak signal unification, we set $p^+$ and $p^-$ in mapping function $m$ to $0.75$ and $0.25$ based on validation.
For  labeling model $p_\phi$, we follow the implementation of Snorkel~\citep{ratner2017snorkel} for efficiency and  train it on CPUs with Adam optimizer.
For noise-aware fine-tuning, we finetune the warmed up  checkpoint with the learning rate of $1e^{-6}$, warmup steps of 500, and the total training steps of 3 epoch. 
We train on 4 NVIDIA Tesla V100 GPUs, and it takes  around only $5000$ steps to reach the best performance.

\section{Results}
The experimental results on TRUE are reported in Table \ref{tab:main_exp}, where we report the performance of our model after warmed up training with NLI as NLI-warmup, and further trained with weak supervision as WeCheck.
Surprisingly, pre-trained language model trained with only NLI-warmup can achieve 80.3 ROC AUC score, which is a comparable performance with previous best metric.
NLI-warmup achieves the second best performance in 5 out of 9 datasets.
After further training with weak supervision, WeCheck improves the evaluation performance over NLI-warmup by 4.5 ROC AUC,  which not only largely surpasses all the baselines but also outperforms previous SOTA metric \textsc{SCzs} by $3.4$ ROC AUC.
Separately on each dataset, WeCheck achieves either the best (6 out of 9) or the second best performance in each dataset.
Specifically, WeCheck achieves 5.4\%, 7.2\%, and 1.6\% of relative improvements over previous best performing methods on summarization, dialogue and paraphrase, respectively.
Furthermore, WeCheck has the lowest variance of 13.2 across different tasks.
This demonstrates that the performance of WeCheck is more comprehensive and general rather than biased towards a certain type of data.
On the MNBM dataset where samples are very different from NLI or QA data (samples in MNBM are sampled from XSUM, where hypothesis are extremely abstractive),  WeCheck largely outperforms previous best metric QAFact by 14.3 point.

\paragraph{11B Baselines} 
We also compare our models with large-scale 11B models based on task transfer.
We compare with two models, $Q^2$ and ANLI based on 11B T5 reported by ~\citet{honovich-etal-2022-true}.
As shown in Table \ref{tab:main_exp}, they surpass the same type of method with smaller parameter size, and can be regarded as approaching the best performance of task transfer based methods can achieve.
However, with only 435M parameters, WeCheck significantly outperforms them by 3-4 points.
This further validates the superiority of our weak supervision learning framework.

\section{Analysis}
To analyse how each module and settings work, we conduct analysis experiments on each module and settings of WeCheck.
\paragraph{Training Mechanism} We first study how the mechanisms proposed in \S\ref{sec:framework} affect the overall framework by removing or replacing  them.
The results are reported in Table~\ref{tab:ablation_mech}.
Most important of all, by removing the NLI-warmup  before weak supervision training, the performance drops significantly on each task and drops an average  of $19.3\%$  on each dataset.
This proves that NLI, as an easier and closely related task, provides a much better initialization for training with weak supervision.
For noise-aware finetuning, we study how filtering potential noisy samples (Eq.~\ref{eq:sample_filter}) and the probabilistic label (Eq.~\ref{eq:soft_label}) affect the overall performance.
After removing noise filtering (w/o Noise Filter in Table~\ref{tab:ablation_mech}), the performance drops around 1-2 points in each task and dataset in average.
By replacing the  probabilistic labels into hard labels (w/ Hard Label in Table~\ref{tab:ablation_mech}), we observe around 0.1-0.2 drops in performance.
This implies how to filter  potential  noisy samples is crucial  in noise aware fine-tuning, and  probabilistic labels also slightly help.

\begin{table}
\renewcommand\arraystretch{1.2}
    \centering
    \small
    \begin{tabular}{l|llll}
    \toprule
          &\textbf{Sum.}&\textbf{Dial.}&\textbf{Para.}&\textbf{Ave} \\
    \midrule
    WeCheck&\textbf{83.0}&\textbf{86.2}&\textbf{89.6}&\textbf{84.8} \\
        \quad \textbf{w/o} NLI-warmup&67.8&75.7&50.7&68.5\\
        \quad \textbf{w/o} Noise Filter&81.6&85.3&78.2&83.7 \\
        \quad \textbf{w/} Hard Label&82.8&86.0&89.5&84.6  \\
    \bottomrule
    \end{tabular}
    \caption{Ablation study of different settings of WeCheck on summarization (Sum.), dialogue (Dial.) and paraphrase (Para.). }
    \label{tab:ablation_mech}
\end{table}

\begin{table}
\renewcommand\arraystretch{1.2}
    \centering
    \small
    \begin{tabular}{lll|llll}
    \toprule
          \textbf{Sum.}&\textbf{Dial.}&\textbf{Para.}&\textbf{Sum.}&\textbf{Dial.}&\textbf{Para.}&\textbf{Ave} \\ 
    \midrule
         &&&77.2&85.4&85.4&80.3\\
        \checkmark&&&\textbf{83.4}&85.2&89.2&84.6\\
         &\checkmark&&72.7&84.2&84.2&77.8\\
         &&\checkmark&77.2&\textbf{86.7}&\textbf{92.1}&81.8\\
         \checkmark&\checkmark&\checkmark&83.0&86.2&89.6&\textbf{84.8}\\

    \bottomrule
    \end{tabular}
    \caption{Analysis on the effects of  different task data. The left block indicates whether using a type of task data  while the right block is the corresponding  performance.}
    \label{tab:task}
\end{table}

\paragraph{Effects of Task} 
We also analyse how each bootstrapped task affect WeCheck.
In Table~\ref{tab:task}, the left block rows indicate whether a type of task samples are used for training, and the right block rows are the corresponding performance.
The first row is the results of NLI-warmup which does not use any task data for training.
The second to forth rows separately train on summarization, dialogue, and paraphrase examples. 
The last row reports the default settings  of WeCheck, which jointly train with all three task samples.
From the results, we can conclude that, joint training on all tasks leads to a better performance on the comprehensive evaluation across tasks.
For single task evaluation except dialogue, training using only the target task examples leads to better performance on this task than joint training.
In horizontal comparisons of single task performance, we observe that summarization examples contribute most to the overall performance, improving the performance of  checking summarization and paraphrase by 6.2 and 3.8 points.
Paraphrase examples benefit evaluating paraphrase and dialogue by  6.7 and 1.3 points.
Dialogue samples worsen the performance of WeCheck.
We suppose that is because these samples are boostrapped from relative weak dialogue models, MemNet and dodecaDialogue, which are not even pre-trained models.
Thus, dialogue samples have no contributions to NLI-warmup.
By contrast, the summarization samples, which are the most difficult type for checking, benefit most to the overall performance.

\paragraph{Computational Efficiency} 
We  analyze the computational efficiency of WeCheck by comparing with other metrics based on different architectures.
As reported in Table~\ref{tab:compu_cost}, we select three other representative metrics:
 SC\textsc{zs} based on sentence-level NLI, FactCC based on document-level NLI, and QAFact based on QA-QG framework.
 All these methods are tested on the TRUE benchmark with a single NVIDIA 32G V100 GPU and we report the relative time cost of each method comparing with WeCheck\footnote{The batch size of each metric is set to the maximum size that the GPU memory can hold.}.
 Despite FactCC is the fastest method reported from the results, its fact checking performance (Table~\ref{tab:main_exp}) is much worse than  others.
Among the rest two methods with  comparable performance, WeCheck is 2.9 times faster than SC\textsc{zs} and 30 times faster than QAFact.

\begin{figure}
    \centering
    \includegraphics[scale=0.19]{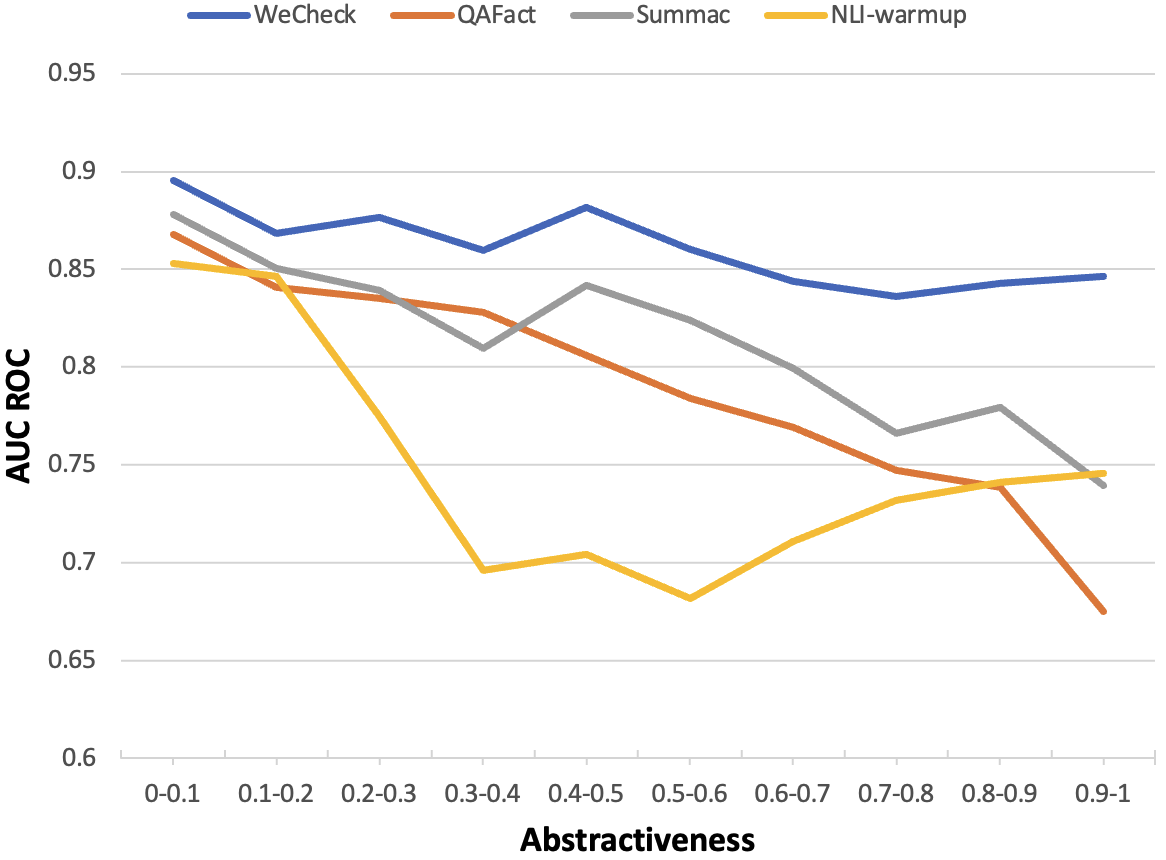}
    \caption{: ROC AUC when splitting TRUE’s data according to abstractiveness.}
    \label{fig:bas_auc}
\end{figure}

\paragraph{Abstractiveness}
As mentioned above, abstractive hypotheses are challenging for current metrics, e.g. XSUM summaries from MNBM.
We give an in-depth analysis of the effect of hypothesis abstractiveness on the metrics performance. 
Following \citet{DBLP:conf/acl/SeeLM17}, we use the percentage of unique unigrams in a hypothesis w.r.t its premise to measure abstractivenss.
Then, we spilt all the examples in TRUE into 10 bins according to their abstractiveness.
For each bin, we measure the ROC AUC of WeCheck and the other three representative baselines: QAFact, Summac, and NLI-warmup.
From the results in Figure \ref{fig:bas_auc}, we observe a significant drop in the performance for all baselines as the hypothesis becomes more abstractive, while, WeCheck keeps its performance (around 0.85). 
Moreover, WeCheck consistently outperforms baseline metrics in every bin of abstractiveness.
This further verifies the superiority of directly training with real task data.

\begin{table}
\renewcommand\arraystretch{1.2}
    \centering
    \small
    \begin{tabular}{rr|rrrr}
    \toprule
&\#size&\textbf{Sum.}&\textbf{Dial.}&\textbf{Para.}&\textbf{Ave} \\ 
    \midrule
         WeCheck&435M&1.0$\times$ &1.0$\times$ &1.0$\times$ &1.0$\times$ \\
          SC\textsc{zs}&59M&3.5$\times$&1.7$\times$&3.4$\times$&2.9$\times$\\
         FactCC&109M&0.2$\times$&0.3$\times$&0.3$\times$&0.2$\times$\\

         QAFact&1097M&24$\times$&26$\times$&75$\times$&30$\times$\\
        
    \bottomrule
    \end{tabular}
    \caption{Inference speed and parameter size (\#size) of different metrics. The right block reports the relative time cost on TRUE  comparing with WeCheck.}
    \label{tab:compu_cost}
\end{table}

\section{Labeling Model} 
We  compare how different data programming based labeling models affect the final metric performance.
In WeCheck, labeling model $p_\phi $ learns to aggregate multi-resource labels to infer the hidden true label.
Comparing concretely, our method is similar to Snorkel~\cite{ratner2017snorkel}.
Because, in our scenario, the number of weak supervision signals is small and their relationships are relatively simple as they are trained from different tasks, we prefer this method over other recent more advanced ones.

In Table~\ref{tab:labeling_model}, we demonstrate the effectiveness of our labeling model by  replacing it with other methods.
In these baselines, simpler methods include:  \textbf{Average Signals}, which simply averages all the weak signals as the probabilistic label $p(y^+)$; \textbf{Major Vote}, which select the most frequently appeared label in a unified weak signal set  as the true label. 
More advanced methods include:  \textbf{Flying Squid}~\cite{DBLP:conf/icml/FuCSHFR20}, which applies an Ising model~\cite{DBLP:journals/ker/Parsons11a} to model more complex relations in a  unified weak signal set; \textbf{Weasel}~\cite{DBLP:conf/nips/CachayBD21} is the current SOTA data programming method, which uses a neural network as the labeling method and trains it end-to-end with the target tasks model; \textbf{DWS}~\cite{parker-yu-2021-named} treats the true label of a sample as the hidden variable and applies Estimation-Maximization (EM) for inference during training. 

From the results in Table~\ref{tab:labeling_model}, our default labeling model outperforms all others.
Furthermore, more complex methods (Flying Squid, Weasel, and EM) perform worse than simpler methods (Ours, Average Signal, and Major Vote).
This further verifies that the relations between weak signals are simple, and complex modeling will not bring further improvements.
From another perspective, overly simplistic approaches without any statistical modeling (Average Signal and Major Vote)  also perform worse than our methods.

\begin{table}
\renewcommand\arraystretch{1.2}
    \centering
    \small
    \begin{tabular}{l|llll}
    \toprule
          Labeling Model&\textbf{Sum.}&\textbf{Dial.}&\textbf{Para.}&\textbf{Ave} \\ 
    \midrule
         Ours&\textbf{83.0}&\textbf{86.2}&\textbf{89.6}&\textbf{84.8} \\

        Average Signal&81.7&86.0&88.7&83.9\\
        Major Vote&81.5&85.6&84.3&83.8\\
        Flying Squid&77.8&84.8&88.4&81.3 \\
        Weasel&74.0&84.4&87.7&79.0\\
        EM&79.0&84.6&86.8&81.7\\
        None&77.2&83.8&85.4&80.3\\
    \bottomrule
    \end{tabular}
    \caption{Performance of WeCheck with different labeling models.}
    \label{tab:labeling_model}
\end{table}

\section{Related Work}
\paragraph{Factual Consistency Evaluation}
Recently, automatically checking factual consistency has become an increasingly popular topic~\cite{https://doi.org/10.48550/arxiv.2203.05227}.
Reasoning over a long range of context for factual evaluation is a challenging task that even human annotators may frequently disagree with each other~\cite{pagnoni-etal-2021-understanding}.
Thus, it is hard to collect a large-scale high-quality dataset for training a fully supervised model, and previous works search for indirect methods.
One branch of them  leverage  the reasoning ability of NLI.
Based on the  model trained on NLI  datasets, e.g. MNLI ~\cite{williams-etal-2018-broad}, ANLI ~\cite{nie-etal-2020-adversarial}, some works aggregate sentence-level entailment score for checking ~\cite{falke-etal-2019-ranking, laban-etal-2022-summac}, while others adopt document-level NLI which directly reasoning  over the full context ~\cite{maynez-etal-2020-faithfulness,gehrmann-etal-2021-gem}.
Another branch of methods apply QA-QG based pipeline for a more fine-grained checking.
QAGS ~\cite{wang-etal-2020-asking} and FEQA~\cite{durmus-etal-2020-feqa} are the earliest attempt on this method, and QuestEval ~\cite{scialom-etal-2021-questeval} and QAFactEval ~\cite{fabbri-etal-2022-qafacteval} further improve this type of methods by applying NLI for answer matching.
\paragraph{Data Programming} In this paper, we mainly focus on data programming~\cite{DBLP:conf/nips/RatnerSWSR16} (DP), a weak supervision paradigm  proposed to infer correct labels based on noisy labels from labeling functions (LFs), which are rule-based decision-making processes that  generate discrete labels.
Following the DP paradigm, Snorkel ~\cite{ratner2017snorkel} is proposed to for rapid training, more recent works study how to adapt label model in DP ~\cite{DBLP:conf/aaai/RatnerHDSPR19,DBLP:journals/corr/abs-2004-06025} or modeling more complex structure between LFs ~\cite{DBLP:conf/icml/FuCSHFR20}.
DP is also applied to several NLP tasks.
DWS ~\cite{parker-yu-2021-named} combine DP and CRF for weakly supervised named entity recognition, ~\citet{DBLP:conf/emnlp/MinCHZ19} apply DP for QA.
Different from all previous tasks, our weak supervision signals are logits from other models, rather than discrete labels generated from rules.

\section{Conclusion}
In this paper, we propose a weakly supervised framework, WeCheck, which aggregates weakly supervised signals from multiple resources and trains a target metric model in a noise-aware manner.
Different from previous metrics that trains from synthetic data or transferred from other tasks, WeCheck directly trains with the real generated text.
WeCheck first annotates each sample with a probabilistic label via a labeling function that aggregates multiple resources.
Then, in the noise-aware finetuning stage, WeCheck applies probabilistic labels to train the target metric model.
Experimental results show that, WeCheck not only surpass previous methods in performance but also time efficient.
Moreover, WeCheck is potential to be compatible with future more stronger metrics, bring further improvements to the overall performance.
\section*{Limitations}
\paragraph{Hyper-parameters Selection}
Some hyper-parameters still acquire careful selection for WeCheck, e.g. $p^+$, $p^-$.
Also, using different set of hyper-parameters  for different tasks and datasets will further boost performance.
Thus, we need to train the model several time and select the best performing parameters based on validation.
\paragraph{End-to-End Training}
WeCheck applies the weak annotation and noise-aware fine-tuning two-step pipeline, where  the noises in the first step will greatly affect the performance of the second step.
By modifying the overall framework into end-to-end training will solve this problem.

\section*{Acknowledgement}
This work was partially supported by 
National Key R\&D Program of China (No. 2022YFC3600402) and National Social Science Foundation Project of China (21\&ZD287).

\bibliography{anthology,custom}
\bibliographystyle{acl_natbib}
\newpage
\appendix

\section{True Benchmark}\label{sec:true_bench}
The TRUE benchmark is composed of the following tasks and datasets.
\paragraph{Abstractive Summarization} \textbf{FRANK} ~\cite{pagnoni-etal-2021-understanding} collect annotations for model-generated summaries on the CNN/DM~\cite{hermann2015teaching} and XSum~\cite{narayan-etal-2018-dont}  datasets, resulting in 2250 annotated system outputs.
\textbf{SummEval (SumE)} ~\cite{fabbri-etal-2021-summeval} collect human judgments for 16 model outputs on 100 articles taken from the CNN/DM dataset.
\textbf{MNBD}~\cite{maynez-etal-2020-faithfulness} sample 500 articles and annotate summaries generated by four different systems on XSum, as well as the gold summaries.
\textbf{QAGS}~\cite{wang-etal-2020-asking} collect 474 generated summaries for CNN/DM and XSum, where each sample is annotated by three annotators.
\paragraph{Dialogue Generation} \textbf{BEGIN}~\cite{DBLP:journals/corr/abs-2105-00071} is a dataset for evaluating the factual consistency of knowledge-grounded dialogue systems.
Dialogue responses are generated by fine-tuning two systems on Wizard of Wikipedia (WoW) ~\cite{DBLP:journals/corr/abs-1811-01241} dataset.
$\mathbf{Q^2}$~\cite{honovich-etal-2021-q2} annotate 1,088  generated dialogue responses from two dialogue models trained on WoW.
\textbf{DialFact (DialF)} ~\cite{gupta-etal-2022-dialfact} introduce a tasks of  dialogue fact-verification   and propose a conversation clams dataset  grounded on Wikipedia.
In TRUE benchmark, one  only need to verify weather a conversation claim is correct given its grounding. 

\paragraph{Paraphrase Detection}
\textbf{PAWS}~\cite{zhang-etal-2019-paws} construct a paraphrase identification with  paraphrase and non-paraphrase pairs from Wikipedia and the Quora Question Pairs (QQP).
In True benchmark, only samples from Wikipedia are applied for verification.



\end{document}